\pdfoutput=1

\documentclass[11pt]{article}
\PassOptionsToPackage{table,xcdraw}{xcolor}
\usepackage[xcdraw]{xcolor}
\usepackage[preprint]{acl}

\usepackage{times}
\usepackage{latexsym}

\usepackage[T1]{fontenc}

\usepackage[utf8]{inputenc}
\usepackage{microtype}
\usepackage{amsmath,amssymb,amsthm}
\usepackage{bbm}
\usepackage{mathabx,mathrsfs}
\usepackage{inconsolata}
\usepackage{xurl}
\usepackage{graphicx}
\usepackage{nicematrix}
\usepackage{booktabs}
\usepackage{natbib}
\usepackage{CJKutf8}
\usepackage{todonotes}
\usepackage{microtype}
\usepackage{pifont}
\usepackage{bm}
\usepackage{graphicx}
\usepackage{booktabs}
\usepackage{tcolorbox}
\usepackage[normalem]{ulem}
\NiceMatrixOptions
       {
         custom-line =
           {
             letter = ; ,
             command = hdashedline ,
             ccommand = cdashedline ,
             tikz = dashed
    } 
}
\definecolor{r-purple}{HTML}{B5739D}
\definecolor{r-blue}{HTML}{0066CC}
\definecolor{r-red}{HTML}{CC0000}
\title{DoCIA: An Online Document-Level Context Incorporation Agent \\ for Speech Translation}

\author{Xinglin Lyu$^{\spadesuit}$, Wei Tang$^{\spadesuit}$, Yuang Li$^{\spadesuit}$, Xiaofeng Zhao$^{\spadesuit}$, Ming Zhu$^{\spadesuit}$, Junhui Li$^{\clubsuit}$, \\ \bf Yunfei Lu$^{\blacklozenge}$,  Min Zhang$^{\spadesuit}$, Daimeng Wei$^{\spadesuit}$, Hao Yang$^{\spadesuit}$, Min Zhang$^{\clubsuit}$\\
$^{\spadesuit}$Huawei Translation Services Center, Beijing, China \\
$^{\blacklozenge}$Huawei Consumer Business Group, Beijing, China \\
$^{\clubsuit}$School of Computer Science and Technology, Soochow University, Suzhou, China \\
\texttt{\{lvxinglin1,tangwei133,zhangming186,yanghao30\}@huawei.com}\\ \texttt{\{lijunhui,minzhang\}@suda.edu.cn}}

\begin{document}
\maketitle
\begin{abstract}
Document-level context is crucial for handling discourse challenges in text-to-text document-level machine translation (MT). Despite the increased discourse challenges introduced by noise from automatic speech recognition (ASR), the integration of document-level context in speech translation (ST) remains insufficiently explored.
In this paper, we develop DoCIA, an online framework that enhances ST performance by incorporating document-level context. DoCIA decomposes the ST pipeline into four stages. Document-level context is integrated into the ASR refinement, MT, and MT refinement stages through auxiliary LLM (large language model)-based modules. Furthermore, DoCIA leverages document-level information in a multi-level manner while minimizing computational overhead. Additionally, a simple yet effective determination mechanism is introduced to prevent hallucinations from excessive refinement, ensuring the reliability of the final results. Experimental results show that DoCIA significantly outperforms traditional ST baselines in both sentence and discourse metrics across four LLMs, demonstrating its effectiveness in improving ST performance.\footnote{Code is available at \url{https://github.com/xllyu-nlp/DoCIA}}
\end{abstract}

\section{Introduction}
\begin{figure}
\centering
\includegraphics[width=1.0\linewidth,trim=20 14 20 12,clip]{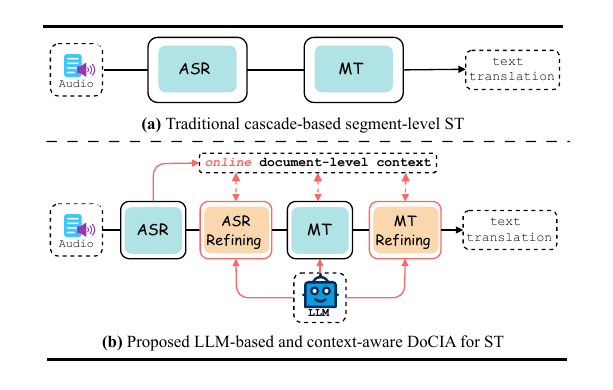}
\caption{The traditional cascade-based ST system (\textit{top}) and our proposed DoCIA for ST (\textit{bottom}). Differently, DoCIA introduces two refinement stages and is LLM-based and context-aware when translating $i$-th audio segment in a speech.}
\label{fig:intro-overview}
\end{figure}
Speech translation (ST) involves translating spoken language into written text in a different language. Despite significant progress in recent years~\cite{zhang_etal_acl_2019_lattice, sperber_etal_acl_2020_speech, ye_etal_interspeech_2021_xstnet, fang_etal_acl_2022_stemm, lei_etal_acl_2023_ckdst}, incorporating document-level context into ST remains a major challenge due to the cross-modal nature of the task. This paper shifts the focus to document-level context\footnote{Also referred to as inter-segment context in ST.} and examines how it can enhance machine translation (MT) when combined with automatic speech recognition (ASR) in the cascaded ST systems.

In a traditional cascaded ST system, as shown in Figure~\ref{fig:intro-overview} (a), the ASR and MT models operate independently at the segment level. This leads to significant discourse-level issues due to the absence of inter-sentence context. These challenges become even more pronounced in ST, where ASR errors—such as misrecognizing entity pronouns or handling disfluencies—further complicate the translation process. Incorporating document-level context offers two key advantages: first, it can potentially correct ASR transcription errors by providing a broader understanding of the context; second, when integrated into the MT model, it helps address discourse phenomena such as entity inconsistencies, coreference resolution, and long-range dependencies~\cite{sennrich_etal_acl_2016, zhang_etal_emnlp_2018, bao_etal_acl_2021_gtrans, bao-etal-2023-targetda, lyu-etal-2024-dempt}. To fully leverage document-level context, we introduce DoCIA—{\bf Do}cument-level {\bf C}ontext {\bf I}ncorporation {\bf A}gent—an online framework specifically designed to improve ST performance by incorporating document-level context.

End-to-end ST systems, which directly translate source-language speech into target-language text, can reduce the propagation of ASR errors. However, these systems suffer from limited interpretability and the challenge of scarce parallel ST data, making it expensive to develop a reliable and effective end-to-end solution. In contrast, the cascading approach—especially with the emergence of powerful large language models (LLMs)~\cite{OpenAI2023GPT4TR, Anil2023PaLM2T, dubey_etal_arxiv_2024_llama3}—provides a more efficient and flexible alternative. The cascading model enables modular optimization in ST, allowing LLMs to be used to enhance performance at various stages of the process. As shown in Figure~\ref{fig:intro-overview} (b), DoCIA takes full advantage of the scalability and flexibility inherent in the cascading approach by breaking the ST process into four key stages: ASR, ASR refinement, MT, and MT refinement. Document-level context is incorporated during the latter three stages (ASR refinement, MT, and MT refinement), improving both transcription and translation through LLM-based agents. Crucially, the document-level context is updated \textit{online} as each segment is processed, ensuring that the context remains current and relevant throughout the translation process.

In addition, we employ two techniques — \textit{a multi-level context integration strategy} and \textit{a refinement determination mechanism} — to enhance the performance of DoCIA. First, while document-level context can be beneficial, it often includes redundant information, with only a small portion being relevant to discourse issues~\cite{kang_etal_emnlp_2020}. Using all available context indiscriminately can even degrade ST performance and increase computational overhead. To address this, we propose a multi-level context integration strategy that retains the advantages of document-level context while reducing redundancy. Second, our two refinement stages are designed to resolve inter-segment inconsistencies using document-level context. In most cases, minimal adjustments are sufficient to address discourse-related issues, as extensive changes may introduce errors such as hallucinations or semantic distortions. To minimize these risks, we introduce a determination mechanism that ensures the refined text remains consistent with the original semantics, improving the output without introducing undesirable changes.

Overall, the main contributions of this paper are summarized as follows:
\begin{itemize}
    \item We extend cascaded ST to four stages and introduce DoCIA, an online agent that enhances ST by progressively incorporating document-level context at each text-to-text stage.
    \item  We propose two techniques to enhance DoCIA: a multi-level document context integration strategy that selectively incorporates context, and a simple determination mechanism to prevent hallucinations during refinement.
    \item We validate DoCIA across five ST directions using four LLMs, including both closed- and open-source models, highlighting the importance of document-level context in ST.
\end{itemize}

\section{DoCIA: Document-level Context Incorporation Agent}
\label{sec:docia}

\begin{figure*}[!t]
    \centering
    \includegraphics[width=1.0\linewidth,trim=20 7 18 7,clip]{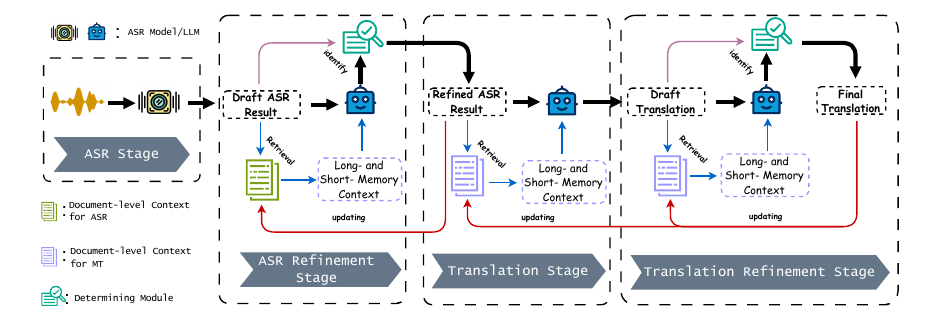}
    \caption{The overall illustration of DoCIA when translating $i$-th audio segment in a speech. The \textcolor{r-blue}{blue}, \textcolor{r-purple}{purple} and \textcolor{r-red}{red} lines denote the \textbf{context retrieving}, \textbf{refinement determining} and \textbf{context updating} processes, respectively.}
    \label{fig:overall-frame}
\end{figure*}
We propose DoCIA, an online agent designed to enhance speech translation (ST) by effectively leveraging document-level context. DoCIA operates in a cascaded four-stage process: ASR, ASR refinement, translation, and translation refinement. Document-level context is incorporated during the ASR refinement, translation, and translation refinement stages (Section \ref{sec:overview-docia}). To optimize context utilization, we introduce a multi-level integration strategy, splitting the context into short- and long-memory components (Section \ref{sec:ml-doc-context}). To prevent hallucinations during refinement, we also propose an effective determination mechanism (Section \ref{sec:ir-methods}).

\subsection{Overview of DoCIA}
\label{sec:overview-docia}

Given a speech $\mathcal{A}=\{a_{1}, a_2, \cdots, a_N\}$, which consists of $N$ audio segments, DoCIA translates these segments sequentially. The overview of DoCIA is illustrated in Figure~\ref{fig:overall-frame}. To explain the translation process, let us consider the $i$-th audio segment $a_i$, as an example. The translation process in DoCIA involves four key stages, which produce the following outputs for $a_i$ : the draft ASR result $\bar{s}_i$, the refined ASR result $s_i$, the draft translation $\bar{t}_i$, and the final refined translation $t_i$. 

\paragraph{ASR Stage.} First, DoCIA generates the draft transcription $\bar{s}_i$ of $a_i$ using an ASR model: 

\begin{equation} 
\label{equ:asr} 
\bar{s}_i = \text{ASR}(a_i). 
\end{equation} 
Here, $\text{ASR}$ refers to the ASR model. Note that in this stage we obtain draft transcription at the segment level.

\paragraph{ASR Refinement Stage.} In this stage, DoCIA aims to correct errors in the draft transcription $\bar{s}_i$ and enhance its quality by incorporating document-level ASR context, denoted as $\mathcal{C}_{asr}=\left(s_1,\cdots,s_{i-1}\right)$. DoCIA uses an LLM to obtain the refined transcription $s_i$ via:
\begin{equation}
\label{equ:asr_refine}
s_i = \text{argmax}~p({s_i|\bar{s_i}, inst_{ar}, C_{asr}, \theta_{llm}}),
\end{equation}
where $inst_{ar}$ represents the instruction for the context-aware ASR refinement task.\footnote{Details of the instructions used in DoCIA can be found in Appendix \ref{sec:prompts-shows}.} $C_{asr} \subseteq \mathcal{C}_{asr}$ is the selected document-level context, which is determined using the strategy described in Section~\ref{sec:ml-doc-context}. The parameter $\theta_{llm}$ refers to the parameters of the LLM.

\paragraph{Translation Stage.} In this stage, DoCIA similarly uses the LLM to translate the transcription $s_i$ while incorporating both the source-side and target-side document-level context $\mathcal{C}_{asr}$ and $\mathcal{T}_{tr}=\left(t_1,\cdots,t_{i-1}\right)$, resulting in the draft translation $\bar{t}_i$ for the audio segment $a_i$. This process is expressed as follows:
\begin{equation} 
\bar{t}_{i} = \text{argmax}~p(\bar{t}_i | s_i, inst_{mt}, C_{tr}, \theta_{llm}),
\label{equ:context-mt} 
\end{equation}
where $inst_{mt}$ represents the instruction for the context-aware translation task. The document-level context for translation, $C_{tr}$, combines the source-side context $C_{asr}$ and the target-side context $T_{tr}$ (also referred to as inter-segment translation history). $T_{tr}$ contains the corresponding refined translations of the segments in $C_{asr}$.

\paragraph{Translation Refinement Stage.} In this stage, we further leverage document-level context to improve the translation through a translation refinement process. Unlike the initial translation stage, where the focus is on generating a translation, the goal here is to enhance the word choice in the draft translation and ensure better cohesion and coherence with the preceding translation history. This process mimics the correction preferences typically applied by human translators. DoCIA again uses the LLM to perform the refinement. Given the draft translation $\bar{t}_i$ for the $i$-th audio segment $a_i$, DoCIA refines it by incorporating document-level context. The refinement process is expressed as follows:
\begin{equation}
\label{equ:mt-refine}
    t_i = \text{argmax}~p(t_i|s_i, inst_{tr}, \bar{t}_i, C_{tr}, \theta_{llm}),
\end{equation}
where $inst_{tr}$ denotes the instruction for the context-aware translation refinement task, $C_{tr}$ is the document-level context, the same as used in Eq.~\ref{equ:context-mt}.  The result, $t_i$, is the final, refined translation of $a_i$.

\subsection{Multi-Level Context Integration}
\label{sec:ml-doc-context}
The translation of different source sentences requires varying amounts of context \cite{kang_etal_emnlp_2020}, and the most relevant context for a given segment should be both dynamic and limited in scope. Therefore, using all preceding transcripts and translations as $C_{asr}$ and $T_{tr}$ may be less effective when translating the $i$-th segment, $a_i$. To address this limitation, we propose a multi-level context, which consists of two components: a short-term context and a long-term context. The multi-level context has a fixed window size $L = m + n$, where $m$ and $n$ represent the number of segments included in the short-term and long-term contexts, respectively.

\paragraph{Short-Memory Context.} Related studies \cite{zhang_etal_emnlp_2018, maruf_etal_naacl_2019} have shown that adjacent sentences are effective in addressing inter-sentence issues during translation. Hence, we define the short-memory context as the $m$ preceding transcript segments of $a_i$ along with their corresponding translations. Specifically, when translating $a_i$, the short-memory context consists of the following: the $m$ preceding transcript segments $C_{asr}^{s} = \{s_{i-m}, \cdots, s_{i-1}\}$ and their corresponding translations ${T_{tr}^{s}} = \{t_{i-m}, \cdots, t_{i-1}\}$.
\paragraph{Long-Memory Context.} Some clues for alleviating inter-segment issues may lie in a longer memory window (i.e., a window size greater than $m$), which makes relying solely on the short-memory context insufficient. To address this, we propose incorporating a long-memory context consisting of $n$ transcript segments and the corresponding translations. More specifically, the transcripts and translation in long-memory context are retrieved all preceding segments, except those already included in the short-memory context:
\begin{equation}
    C_{asr}^{l} = f(q_i, \mathcal{C}, n),
\end{equation}
where $\mathcal{C} = \{s_1, \cdots, s_{i-m}\}$ represents the set of transcripts preceding the short-memory context and $f$ is a retrieval function. Given a query $q_i$, $f$ returns the top $n$ matching transcript segments from $\mathcal{C}$, forming $C_{asr}^{l}$. During ASR refinement, $q_i$ is set to $\bar{s}_i$ while during translation and translation refinement, $q_i$ is set to ${s_i}$. Once we obtain $C_{asr}^l$, we can easily retrieve the corresponding translations $T_{tr}^{l}$ from $\mathcal{T}_{tr}$. In the strategy, we use BM25~\cite{bm25s} as the retrieval function $f$. Finally, the document-level context used in Eq. \ref{equ:asr_refine},\ref{equ:context-mt} and \ref{equ:mt-refine} combines long- and short-memory context:
\begin{equation}
    C_{asr} = C_{asr}^{l} + C_{asr}^s,
\end{equation}
\begin{equation}
    T_{tr} = T_{tr}^{l} + T_{tr}^s.
\end{equation}

\subsection{Refinement Determination Mechanism}
\label{sec:ir-methods}

\begin{table*}[!t]
\setlength{\tabcolsep}{3.85pt}
\renewcommand{\arraystretch}{0.95}
\centering
\resizebox{\textwidth}{!}{\begin{NiceTabular}{l|cccccccccccc}
\toprule
\Block[l]{2-1}{\textbf{System}} & \Block[c]{1-2}{\textbf{En $\Rightarrow$ De}} &  & \Block[c]{1-2}{\textbf{En $\Rightarrow$ It}} & & \Block[c]{1-2}{\textbf{En $\Rightarrow$ Pt}} & & \Block[c]{1-2}{\textbf{En $\Rightarrow$ Ru}} & & \Block[c]{1-2}{\textbf{En $\Rightarrow$ Ro}} && \Block[c]{1-2}{\textbf{\textit{Average}}} &\\
\cmidrule(lr){2-3}\cmidrule(lr){4-5}\cmidrule(lr){6-7}\cmidrule(lr){8-9}\cmidrule(lr){10-11}\cmidrule(lr){12-13}
& $s$-Comet & $d$-Comet & $s$-Comet & $d$-Comet & $s$-Comet & $d$-Comet & $s$-Comet & $d$-Comet & $s$-Comet & $d$-Comet& $s$-Comet & $d$-Comet\\

\rowcolor[gray]{0.8}
\Block[c]{1-13}{\texttt{LLaMA-3.1-8B}}\\
ASR-SMT                             &78.01      &5.680      &79.67      &5.619      &80.57      &5.438      &76.36     &5.168      &79.07       &5.372&78.73&5.455\\
ASR-DMT                             &77.88      &5.712      &79.79      &5.651      &80.69      &5.477      &76.99     &5.211      &79.01      &5.401&78.87&5.490\\
\hdashedline
$\text{DoCIA}_{a}$                  &78.11      &5.764      &80.03      &5.703      &81.45      &5.519      &77.16     &5.288      &79.69      &5.473&79.29&5.549\\
$\text{DoCIA}_{a\text{-}m}$         &78.50      &5.801      &80.53      &5.792      &\cellcolor{orange!16.8}\underline{\bf 81.99}&5.621      &78.03     &5.401      &80.39      &5.599&79.89&5.643\\
$\text{DoCIA}_{a\text{-}m\text{-}p}$&\cellcolor{orange!11.4}\underline{\bf 79.15}&\cellcolor{blue!13}\underline{\bf 5.912}&\cellcolor{orange!12.8}\underline{\bf 80.88}&\cellcolor{blue!15}\underline{\bf 5.909}&81.75      &\cellcolor{blue!18}\underline{\bf 5.757}&\cellcolor{orange!32.8}\underline{\bf78.39}&\cellcolor{blue!22}\underline{\bf 5.556}&\cellcolor{orange!21.8}\underline{\bf 80.54}&\cellcolor{blue!22}\underline{\bf 5.734}&\cellcolor{orange!20}\underline{\bf80.15}&\cellcolor{blue!20}\underline{\bf 5.774}\\

\rowcolor[gray]{0.8}
\Block[c]{1-13}{\texttt{LLaMA-3.1-70B}}\\
ASR-SMT                             &81.11      &5.997       &82.01      &5.811      &82.03      &5.626     &80.26      &5.686      &83.28      &5.808 &81.73&5.785\\
ASR-DMT                             &81.54      &6.143       &82.36      &5.976      &82.85      &5.745     &80.99      &5.867      &83.15      &5.979&82.17&5.942\\
\hdashedline
$\text{DoCIA}_{a}$                  &81.64      &6.098       &82.55      &5.948      &82.53      &5.740     &81.26      &5.803      &83.82      &5.935&82.36&5.905\\
$\text{DoCIA}_{a\text{-}m}$         &\cellcolor{orange!19.8}\underline{\bf82.69} &6.155       &\cellcolor{orange!21.8}\underline{\bf 83.85}&6.132      &83.87      &5.893     &\cellcolor{orange!50.8}\underline{\bf 82.73}&6.034      &84.64      &6.131&83.57&6.069\\
$\text{DoCIA}_{a\text{-}m\text{-}p}$&82.63      &\cellcolor{blue!21}\underline{\bf 6.373} &83.66      &\cellcolor{blue!30}\underline{\bf 6.264}&\cellcolor{orange!32.8}\underline{\bf 83.99}&\cellcolor{blue!25}\underline{\bf6.037}&82.69      &\cellcolor{blue!26}\underline{\bf 6.168}&\cellcolor{orange!41}\underline{\bf 85.32}&\cellcolor{blue!35}\underline{\bf 6.365}&\cellcolor{orange!38}\underline{\bf 83.66}&\cellcolor{blue!30}\underline{\bf 6.241}\\

\rowcolor[gray]{0.8}
\Block[c]{1-13}{\texttt{GPT-4o-mini}}\\
ASR-SMT                             & 82.01      &6.001     &83.14     &5.683     &82.56     &5.671     &82.21      &5.827      &84.25     &5.940 &82.83 &5.824\\
ASR-DMT                             & 82.42      &6.108     &83.52     &5.833     &83.32     &5.943     &82.80      &5.948      &84.82     &6.018&83.37&5.970\\
\hdashedline
$\text{DoCIA}_{a}$                  & 82.99      &6.174     &83.70     &6.004     &84.03     &5.804     &82.77      &5.935      &84.89     &6.082&83.68&6.000\\
$\text{DoCIA}_{a\text{-}m}$         &\cellcolor{orange!21.8}\underline{\bf 83.75}&6.366     &84.54     &6.233     &\cellcolor{orange!32.8}\underline{\bf84.57}&6.024     &84.10      &6.215      &85.46     &6.213&84.48&6.210\\
$\text{DoCIA}_{a\text{-}m\text{-}p}$& 83.64      &\cellcolor{blue!27}\underline{\bf6.444}&\cellcolor{orange!20.8}\underline{\bf84.76}&\cellcolor{blue!46}\underline{\bf6.387}&84.51     &\cellcolor{blue!40}\underline{\bf6.297}&\cellcolor{orange!35.8}\underline{\bf84.32} &\cellcolor{blue!24}\underline{\bf6.286} &\cellcolor{orange!42}\underline{\bf86.34}&\cellcolor{blue!29}\underline{\bf6.424}&\cellcolor{orange!38}\underline{\bf 84.71}&\cellcolor{blue!34}\underline{\bf 6.368}\\

\rowcolor[gray]{0.8}
\Block[c]{1-13}{\texttt{GPT-3.5-turbo}}\\
ASR-SMT                             &81.51     &5.974     &81.74      &5.732      &82.40      &5.658      &79.21      &5.566       &82.91      &5.644&81.55&5.715\\
ASR-DMT                             &81.68     &5.977     &81.93      &5.760      &82.53      &5.687      &79.50      &5.611       &83.30      &5.687&81.78&5.744\\
\hdashedline
$\text{DoCIA}_{a}$                  &81.70     &5.961     &82.30      &5.705      &82.63      &5.634      &79.57      &5.651       &83.77      &5.601&81.99&5.710\\
$\text{DoCIA}_{a\text{-}m}$         &82.93     &6.126     &83.18      &5.838      &83.60      &5.763      &81.71      &5.804       &84.68      &5.891& 83.22& 5.884\\
$\text{DoCIA}_{a\text{-}m\text{-}p}$&\cellcolor{orange!18.8}\underline{\bf82.95}&\cellcolor{blue!13}\underline{\bf6.192}&\cellcolor{orange!21.8}\underline{\bf 83.39}&\cellcolor{blue!17}\underline{\bf 5.997}&\cellcolor{orange!20.8}\underline{\bf 83.90}&\cellcolor{blue!10}\underline{\bf 5.797}&\cellcolor{orange!53.8}\underline{\bf 81.97}&\cellcolor{blue!15}\underline{\bf 5.841} &\cellcolor{orange!41}\underline{\bf 85.01}&\cellcolor{blue!25}\underline{\bf 6.033}&\cellcolor{orange!38}\underline{\bf 83.45}&\cellcolor{blue!17}\underline{\bf 5.973}\\
\bottomrule
\end{NiceTabular}}
\caption{$s$-Comet and $d$-Comet scores on five ST directions when using various LLMs. The column of \textbf{\textit {Average}} refers to the averaged performance across all translation directions. The top score in each block is highlighted in \underline{\bf bold} font. Darker colors indicate greater improvements.}
\label{tab:comet-eval}
\end{table*}

To enhance both the overall quality of transcriptions and translations, DoCIA incorporates two context-aware refinement processes. These processes aim to leverage document-level context, improving the coherence and cohesion between segments. Given that inter-segment issues are typically sparse, the refinement process generally focuses on making minor adjustments to the source input. However, excessive refinement could introduce errors that distort the original meaning, leading to hallucinations \cite{Xu2024HallucinationII}. To address this, we introduce a refinement determination mechanism. Specifically, we define a refinement threshold: if the percentage of modifications in the refined output exceeds this threshold, the refinement is discarded, and the original input is retained as the final output:

\begin{equation}
R = \begin{cases} 
O & \text{if } g(O, I) \geq \lambda \\
I & \text{if } g(O, I) < \lambda
\end{cases}
\end{equation}
where $I$ denotes the original input (i.e., the draft text ($\bar{s}_i$ or $\bar{t}_i$), $O$ is the refined output, and $R$ is the final output. $\lambda$ denotes the threshold of modification. We use the normalized \textit{indel similarity} between $I$ and $O$ as $g$:
\begin{equation}
\label{equ:idt-f}
    g(O, I) = 1 - \frac{d(O, I)}{|I| + |O|},
\end{equation}
where $d(\cdot)$ is \textit{Levenshtein edit distance} function, $|\cdot|$ denotes segment length.

For simplicity, we use the same threshold $\lambda$ for both ASR and translation refinement.

\section{Experimentation}
In this section, we validate the effectiveness of DoCIA on five ST transation tasks.
\subsection{Experimental Settings}
\paragraph{Datasets.} We conduct our experiments on the MuST-C test sets \cite{di_gangi_etal_naacl_2019_mustc}, which are extracted from TED talks and consist of document-level and sentence/segment-level parallel corpora. In our study, we focus on five language pairs: English (En) $\Rightarrow$ \{German (De), Italian(It), Portuguese (Pt), Romanian (Ro), Russian (Ru)\}. Each test set contains approximately 2.5K segments drawn from 27 talks (documents). 
\paragraph{Metrics.} We evaluate translation quality using two COMET-based metrics. For segment-level evaluation, we use s-COMET with the \texttt{wmt22-comet-da} model \cite{rei-etal-2020-comet}. For document-level evaluation, we use d-COMET with the \texttt{wmt21-comet-qe-mqm} model \cite{easy_doc_mt}, which incorporates document-level context to assess improvements across segments. Additionally, inspired by recent advances in LLM-based evaluation methods \cite{zheng2023judging}, we also leverage \texttt{GPT-4o} for translation assessment (see Appendix \ref{sec:llm-base-eval-all} for details).

\paragraph{Models and Hyperparameters.} DoCIA is built upon four LLMs: two closed-source models, \texttt{GPT-4o-mini} and \texttt{GPT-3.5-turbo} \cite{OpenAI2023GPT4TR}, and two open-source models, \texttt{LLaMA-3.1-8B} and \texttt{LLaMA-3.1-70B} \cite{dubey_etal_arxiv_2024_llama3}, and run inference of open-source models with 8$\times$ Ascend 910B NPUs. For all experiments, we use \texttt{Whisper-medium} \cite{radford_etal_icml_2023_whisple} to generate draft ASR results. A discussion on the impact of different ASR models is provided in Appendix \ref{sec:asr-effect}. We set the context window size $L$ as 6, with $m=n=3$. The refinement threshold $\lambda$ is set to 0.7. Further model and hyperparameter selection details are discussed in Appendix \ref{sec:asr-effect} and \ref{sec:lamda-func}.

\paragraph{Comparison System.} We implement the following two systems for comparison: 1) {\bf ASR-SMT}, which performs segment-level translation directly on the draft ASR output; 2) {\bf ASR-DMT}, which performs context-aware translation directly on the draft ASR output, using all preceding ASR segments to incorporate document-level context. To better analyze the impact of document-level context at different stages, we define three configurations of DoCIA: 1) $\textbf{DoCIA}_{a}$, which only the context-aware ASR refinement stage; $\textbf{DoCIA}_{a\text{-}m}$, which integrates both context-aware ASR refinement and MT; and $\textbf{DoCIA}_{a\text{-}m\text{-}p}$, which in all three text-to-text stages, leverages document-level information.

\subsection{Main Results}
We report our main results in Table \ref{tab:comet-eval} and LLM-based evaluation in Appendix \ref{sec:llm-base-eval-all}. Additionally, we report the ASR refinement results in Appendix \ref{sec:asr-refine-res}.
From them, we have the following observations: 
\paragraph{DoCIA gains a great improvement over baseline systems.} DoCIA delivers substantial gains over both ASR-SMT and ASR-DMT, particularly in $d$-Comet scores, highlighting its effectiveness in handling document-level context.
For example, with the \texttt{LLaMA-3.1-8B} model, the configuration $\text{DoCIA}_{a\text{-}m\text{-}p}$ (which fully integrates document-level context) achieves an average $s$-Comet score of 80.15 and a $d$-Comet score of 5.774. This outperforms both ASR-SMT and ASR-DMT, with improvements of +1.42 in $s$-Comet and +0.319 in $d$-Comet over ASR-SMT. Similarly, with the \texttt{GPT-4o-mini} model, $\text{DoCIA}_{a\text{-}m\text{-}p}$ shows even more pronounced improvements, surpassing ASR-SMT by +1.88 in $s$-Comet and +0.544 in $d$-Comet. This demonstrates the effectiveness of incorporating document-level context in ST. 
\paragraph{Better base model brings more significant improvement.} DoCIA yields more substantial improvements when applied to a better base model such as \texttt{LLaMA-3.1-70B} and \texttt{GPT-4o-mini}. For instance, with \texttt{LLaMA-3.1-8B}, DoCIA results in improvements of +1.42 in $s$-Comet and +0.319 in $d$-Comet on average, compared to ASR-SMT. While using \texttt{GPT-4o-mini} as the base model, DoCIA achieves even larger gains, with improvements of +1.93 in $s$-Comet and +0.466 in $d$-Comet. This may suggest that more powerful LLMs can better utilize document-level context within the DoCIA framework, resulting in improved speech translation quality and enhanced context.
\paragraph{Document-level context boosts performance more when combined with other stages than using alone.} When the document-level context is integrated into the ASR refinement phase alone (i.e., DoCIA$_{a}$), the improvements in $s$-Comet and $d$-Comet scores are relatively small but still noticeable. For example, with \texttt{LLaMA-3.1-8B}, DoCIA$_{a}$ shows a modest improvement of +0.56 in $s$-Comet and +0.094 in $d$-Comet on average compared to ASR-SMT. However, the performance boost becomes much more substantial when combined with additional stages. For example, compared to DoCIA$_{a}$ which solely incorporates document-level context during ASR refinement, $\text{DoCIA}_{a\text{-}m}$ bring a + 1.12 $s$-Comet and + 0.164 $d$-Comet gains. This demonstrates that the multi-stage integration approach effectively unlocks the potential of document-level context, enabling comprehensive optimization of ST.

\section{Discussion and Analysis}
In this section, we use the En $ \Rightarrow $ De and En $ \Rightarrow $ Ru tasks, with \texttt{LLaMA-3.1-8B} and \texttt{GPT-4o-mini}, as representative examples to explore how DoCIA enhances ST performance. 
\subsection{Ablation Study}
\label{sec:abl-study}
\begin{table}[!t]
    \centering
    \resizebox{\linewidth}{!}{
    \begin{NiceTabular}{c|cccc}
    \toprule
    \Block[l]{2-1}{\textbf{System}} & \Block[c]{1-2}{\textbf{En $\Rightarrow$ De}} &  & \Block[c]{1-2}{\textbf{En $\Rightarrow$ Ru}}\\
    
    \cmidrule(lr){2-3}\cmidrule(lr){4-5}
    & $s$-Comet & $d$-Comet & $s$-Comet & $d$-Comet \\
    
    \rowcolor[gray]{0.8}
    \Block[c]{1-5}{\texttt{LLaMA-3.1-8B}}\\
    
    DoCIA                    & 79.15 & 5.912 & 78.39 & 5.556\\
    \hdashedline
    ~~~~~~\textit{w/o} R.D.   & 78.33 & 5.812 & 77.50 & 5.431\\
    ~~~~~~\textit{w/o} S.C. & 78.63 & 5.792 & 77.81 & 5.331 \\
    ~~~~~~\textit{w/o} L.C. & 78.41 & 5.761 & 77.88 & 5.311\\
    
    \rowcolor[gray]{0.8}
    \Block[c]{1-5}{\texttt{GPT-4o-mini}}\\
     DoCIA                   & 83.64 & 6.444 & 84.32 & 6.286\\
    \hdashedline
    ~~~~~~\textit{w/o} R.D. & 82.66 & 6.299 & 83.11 & 6.116\\
    ~~~~~~\textit{w/o} S.C. & 83.11 & 6.231 & 83.88 & 6.061\\
    ~~~~~~\textit{w/o} L.C. & 83.01 & 6.201 & 83.77 & 6.011\\
    \bottomrule
    \end{NiceTabular}}
   
\caption{Ablation study for \textit{refinement determination} (R.D.) and \textit{multi-level context integration}. \textit{w/o S.C.} disables short-memory context, using only the top $L$ matching segments from the long-memory context. \textit{w/o L.C.} disables long-memory context and uses the $L$ preceding segments from short-memory context instead.}
\label{tab:abl-id-mc}
\end{table}
In this section, we conduct an ablation study to evaluate the contributions and impacts of individual components within DoCIA, including the multi-level context integration and the refinement determination. As shown in Table \ref{tab:abl-id-mc}, the comparison shows that the refinement determination (\textit{w/o} R.D.) primarily affects $s$-Comet, while the multi-level context integration influences $d$-Comet more. For instance, removing the refinement determination module leads to a 0.98 drop in $s$-Comet and 0.145 in $d$-Comet for En$\Rightarrow$De translation using the \texttt{GPT-4o-mini} model. While disabling the long-memory context in multi-level context integration (\textit{w/o} L.C.) causes a decrease of 0.63 in $s$-Comet and 0.243 in $d$-Comet. This suggests that the two components are complementary, highlighting the necessity of their combined use. Furthermore, we observe that long-memory context has a more substantial effect on performance than short-term context, underscoring the importance of leveraging long-range dependency.

\subsection{Multi-Dimension Evaluation via GPT-4o}
\begin{table}[!t]
    \centering
    \resizebox{\linewidth}{!}{
    \begin{NiceTabular}{l|crrcrr}
    \toprule
    \Block[l]{2-1}{\textbf{System}} & \Block[c]{1-3}{\textbf{En $\Rightarrow$ De}} & & & \Block[c]{1-3}{\textbf{En $\Rightarrow$ Ru}}\\
    
    \cmidrule(lr){2-4}\cmidrule(lr){5-7}
    & Fluency & LE$\downarrow$ & GE$\downarrow$ & Fluency & LE$\downarrow$ & GE$\downarrow$\\
    
    \rowcolor[gray]{0.8}
    \Block[c]{1-7}{\texttt{LLaMA-3.1-8B}}\\
    
    ASR-SMT & 3.01       & 5.21      & 4.28       & 2.89     &6.11      & 4.75\\
    ASR-DMT & 3.11       & 4.32      & 3.63       & 3.12     &5.28      & 4.63\\
    DoCIA   &\underline{\bf 3.76} &\underline{\bf1.98} &\underline{\bf1.42}  &\underline{\bf3.71}&\underline{\bf3.32} &\underline{\bf2.29}\\
    
    \rowcolor[gray]{0.8}
    \Block[c]{1-7}{\texttt{GPT-4o-mini}}\\
    ASR-SMT & 4.35      & 3.21      & 2.28       & 4.01     &3.78      & 2.75\\
    ASR-DMT & 4.47      & 2.01      & 1.77       & 4.24     &2.61      &1.63\\
    DoCIA   &\underline{\bf5.16} &\underline{\bf1.01} &\underline{\bf0.79}  &\underline{\bf4.98}&\underline{\bf1.33} &\underline{\bf0.82}\\
    \bottomrule
    \end{NiceTabular}}
    \caption{Evaluation results on test set by \texttt{GPT-4o}.}
\label{tab:eval-fine-gain}
\end{table}
\label{sec:fine-eval-by-llm}
In Appendix \ref{sec:llm-base-eval-all}, we compared overall win rates via \texttt{GPT-4o}-based metric across various systems. In this section, we extend the evaluation by using \texttt{GPT-4o} to assess various discourse phenomena. Specifically, we follow \citet{sun2024instruction} and ask \texttt{GPT-4o} to evaluate the inter-sentence fluency, lexical cohesion errors (LE), and grammatical cohesion errors (GE) in the given translations, using reference translations for comparison. As shown in Table \ref{tab:eval-fine-gain}, ASR-DMT outperforms ASR-SMT, demonstrating that integrating inter-segment context significantly reduces lexical and grammatical cohesion errors while improving overall fluency. Notably, DoCIA achieves the best performance on all translation tasks across all three metrics, further highlighting its effectiveness in leveraging inter-segment context. Additional discourse metrics for translation quality evaluation are provided in Appendix \ref{sec:apt-eval} and \ref{sec:ltcr-eval}.  

\subsection{Effect of \textit{Online/Offline} Setting}
\begin{table}[!t]
    \centering
    \resizebox{\linewidth}{!}{
    \begin{NiceTabular}{c|cccc}
    \toprule
    \Block[l]{2-1}{\textbf{System}} & \Block[c]{1-2}{\textbf{En $\Rightarrow$ De}} &  & \Block[c]{1-2}{\textbf{En $\Rightarrow$ Ru}}\\
    
    \cmidrule(lr){2-3}\cmidrule(lr){4-5}
    & $s$-Comet & $d$-Comet & $s$-Comet & $d$-Comet \\
    
    \rowcolor[gray]{0.8}
    \Block[c]{1-5}{\texttt{LLaMA-3.1-8B}}\\
    
    DoCIA                    & 79.15 & 5.912 & 78.39 & 5.556\\
    \hdashedline
    ~~~~\textit{w/ {\bf \textit{offline}}}    & 78.24 & 5.783 & 77.30 & 5.342\\
    
    \rowcolor[gray]{0.8}
    \Block[c]{1-5}{\texttt{GPT-4o-mini}}\\
     DoCIA                   & 83.64 & 6.444 & 84.32 & 6.286\\
    \hdashedline
    ~~~~\textit{w/ {\bf \textit{offline}}}   & 82.81 & 6.252 & 83.01 & 6.095\\
    \bottomrule
    \end{NiceTabular}}
   
    \caption{Performance comparison between \textit{online} and \textit{offline} DoCIA on test set. }
\label{tab:offline-online}
\end{table}
In DoCIA, the document context is updated in real-time during the translation process, following an \textit{online} fashion. In contrast, we also compare this with the \textit{offline} setting, where the context is not updated and relies solely on segment-level translation/ASR results, denoted as \textit{offline} DoCIA. As shown in Table \ref{tab:offline-online}, the \textit{offline} DoCIA shows a significant drop in performance compared to \texttt{online} DoCIA. For example, in the En$\Rightarrow$Ru task using the \texttt{LLaMA-3.1-8B} model, \textit{offline} DoCIA results in a -1.09 decrease in $s$-Comet score and a -0.214 decrease in $d$-Comet score. This suggests that DoCIA's performance is highly sensitive to the quality of the context, with real-time updates leading to more accurate and effective context, which in turn significantly improves speech translation quality.

\subsection{Human Evaluation}
\begin{table}[!t]
    \centering
    \resizebox{\linewidth}{!}{
    \begin{NiceTabular}{l|crrcrr}
    \toprule
    \Block[l]{2-1}{\textbf{System}} & \Block[c]{1-3}{\textbf{En $\Rightarrow$ De}} & & & \Block[c]{1-3}{\textbf{En $\Rightarrow$ Ru}}\\
    
    \cmidrule(lr){2-4}\cmidrule(lr){5-7}
    & DA & CE$\downarrow$ & CTE$\downarrow$ & DA & CE$\downarrow$ & CTE$\downarrow$\\
    
    \rowcolor[gray]{0.8}
    \Block[c]{1-7}{\texttt{LLaMA-3.1-8B}}\\
    
    ASR-SMT & 89.7       & 13.0      & 16.0       & 76.7  &16.3      & 20.3\\
    ASR-DMT & 90.1       & 9.5       & 14.0       & 78.0  &11.5      & 17.3\\
    DoCIA   &\underline{\bf 92.3} &\underline{\bf5.5} &\underline{\bf7.5}  &\underline{\bf80.7}&\underline{\bf8.5} &\underline{\bf 11.5}\\
    
    \rowcolor[gray]{0.8}
    \Block[c]{1-7}{\texttt{GPT-4o-mini}}\\
    ASR-SMT & 92.5      & 8.0      & 12.0       & 81.3     &12.3  & 15.0\\
    ASR-DMT & 92.8      & 7.3      & 13.0       & 82.6     &11.1  &12.3\\
    DoCIA   &\underline{\bf94.7} &\underline{\bf3.3} &\underline{\bf6.0}  &\underline{\bf85.0}&\underline{\bf7.3} &\underline{\bf9.5}\\
    \bottomrule
    \end{NiceTabular}}
    \caption{Results of human evaluation on the test set.}
\label{tab:huaman-eval}
\end{table}
We use the Direct Assessment (DA) \cite{graham-etal-2017-da} to evaluate the translation quality of DoCIA and its counterparts. Here, human evaluators compare machine translations with human-produced references in the same language and assign a score from 1 to 100. For each translation direction, we randomly select 4 talks, totaling 312 audio segments, and have two professional translators score the translations from DoCIA, ASR-SMT, and ASR-DMT. Additionally, we report the average counts (per talk) of coherence errors (CE) and content translation errors (CTE) annotated by evaluators. The results, presented in Table \ref{tab:huaman-eval}, show that DoCIA outperforms the others with higher DA scores and fewer CE and CTE scores, providing strong evidence of its effectiveness. For more details of human evaluation, refer to Appendix \ref{sec:details-da}.
\subsection{Effect of Context Window}
\label{sec:ctx-wins}

\begin{figure*}[!t]
\centering
\includegraphics[width=1.0\linewidth,trim=18 25 16 23,clip]{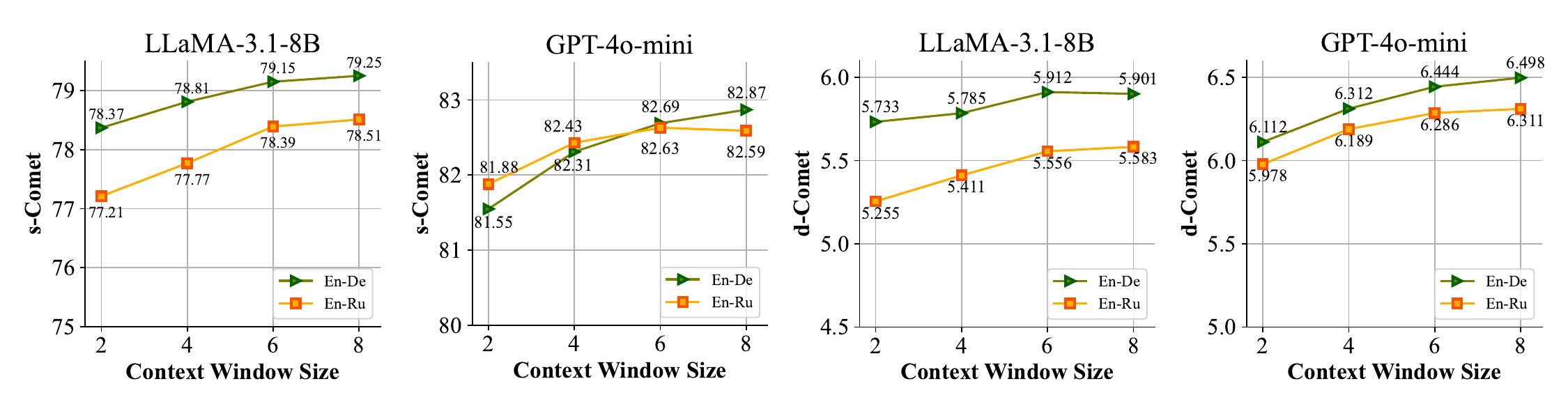}
\caption{Performance comparison when setting different context window size $L$.}
\label{fig:wind-size-comp}
\end{figure*}
\begin{figure*}[!t]
\centering
\includegraphics[width=1.0\linewidth,trim=22 6 20 4,clip]{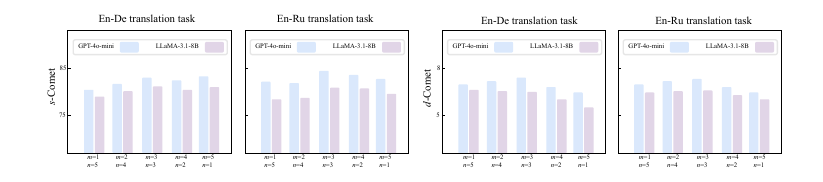}
\caption{Performance comparison when setting different combinations of $m$ and $n$.}
\label{fig:m-n-comp}
\end{figure*}

In this section, we examine the impact of the context window from two perspectives: 1) varying the context window size $L$, and 2) exploring different combinations of $m$ and $n$ while keeping $L$ fixed. As shown in Figure \ref{fig:wind-size-comp}, increasing the context window size $L$ generally improves performance across all metrics. However, the gains start to diminish when $L$ exceeds 6. Figure \ref{fig:m-n-comp} illustrates the effects of different $m$ and $n$ combinations. Similar to the trends observed in Section \ref{sec:abl-study}, we find that reducing the short-memory context (i.e., smaller $m$) has a more significant impact on $s$-Comet, while decreasing the long-memory context (i.e., smaller $n$) affects the $d$-Comet score more. This further reinforces the complementary nature of short- and long-memory contexts in DoCIA. 

\section{Related Work}
\paragraph{LLM-based Autonomous Agents.} 
LLM-based autonomous agents have recently demonstrated impressive capabilities across a variety of natural language processing tasks. For long-context comprehension and processing, researchers such as \citet{park2023generative}, \citet{wang2023enhancing}, and \citet{lee2024human} have developed specialized memory and retrieval mechanisms. In efforts to improve output quality, \citet{xu2024llmrefine}, \citet{wang-etal-2024-taste}, and \citet{feng2024improving} have employed prompting techniques that allow LLMs to self-assess and refine their results. Additionally, \citet{li2023camel}, \citet{liang2023encouraging}, \citet{li2024agent}, \citet{wu2024perhaps} and \citet{Wang2024DelTAAO} boost LLM performance on specific tasks through multi-agent collaboration.
\paragraph{Speech-to-Text Translation.} 
Existing studies on ST can be roughly categorized into two groups: cascade-based and end-to-end approaches. The cascade-based system ~\cite{zhang_etal_acl_2019_lattice,sperber_etal_acl_2020_speech,lam_etal_icassp_2021_cascaded} separates ASR and text translation stages, which doesn’t require parallel audio-translation data and can fully leverage ASR and text translation corpus for ST. While the end-to-end system combines these stages and is trained on parallel audio-translation data using strategies such as multi-task learning~\cite{ye_etal_interspeech_2021_xstnet}, contrastive learning~\cite{ye_etal_naacl_2022_const,zhang_etal_emnlp_2022_fcgcl,ouyang_etal_acl_2023_waco}, sequence mixup~\cite{fang_etal_acl_2022_stemm,yin_etal_emnlp_2023_improving,zhang_etal_acl_2023_simple,zhou_etal_acl_2023_cmot}, knowledge distillation~\cite{tang_etal_acl_2021_improving,lei_etal_acl_2023_ckdst}, regularization~\cite{han_etal_acl_2023_modality,gao_etal_naacl_2024_empirical}, pre-training~\cite{wang_etal_acl_2020_curriculum,alinejad_sarkar_acl_2020_effectively,tang_etal_acl_2022_unified,zhang_etal_emnlp_2022_speechut}, and data augmentation~\cite{pino_etal_iwslt_2019_harnessing,pino_etal_interspeech_2020_self_training,lam_etal_acl_2022_sample}. Recently, with the rise of LLMs, some research has explored combining speech encoders with LLMs for end-to-end ST~\cite{wu_etal_asru_2023_decoder,chen_etal_icassp_2024_salm}. However, few studies explore the effect of document-level information in the ST.

\paragraph{Document-Level Text Translation.} Document-level context has already been widely considered in text translation studies whether based on the lightweight neural machine translation models \cite{jean_etal_2017,wang_etal_emnlp_2017,voita_etal_acl_2018,maruf_etal_naacl_2019,kang_etal_emnlp_2020,bao_etal_acl_2021_gtrans,sun_etal_2022_acl_rethinking,bao-etal-2023-targetda} or powerful LLMs \citet{wang-etal-2023-document-level, wu-hu-2023-exploring, Wu2024AdaptingLL, Li2024Enhancing, koneru_etal_2024_contextual,lyu-etal-2024-dempt}. These studies primarily focus on efficiently leveraging document-level context to address inter-sentence translation issues. For example, \citet{lyu_etal_2021_emnlp_ltcr} proposed to model the information of the repeated words in a document to alleviate their translation inconsistencies. \citet{Wu2024AdaptingLL} explored the effective tuning methods to make LLMs harness the benefits brought by document-level context. Despite the effectiveness of document-level context in text translation, it remains underexplored in ST.

\section{Conclusion}
Inspired by the success of incorporating document-level context in text-to-text MT, we propose DoCIA, an online LLM-based agent designed to improve ST performance by integrating document-level context. DoCIA breaks the whole ST process into four stages, producing the final translation in a cascading manner. Additionally, we introduce a multi-level context integration strategy and a refinement determination mechanism to enhance DoCIA’s ability to utilize inter-segment context while minimizing hallucinations during refinement. Experimental results across five ST tasks, using four different LLMs, demonstrate that DoCIA effectively addresses discourse issues from both the ASR and MT stages, leading to significant improvements in overall ST quality.

\section*{Limitations}
In this paper, we propose a document-level context incorporation agent for ST, focusing primarily on its effectiveness in improving ST performance rather than optimizing inference speed. The inference requires multiple calls to LLMs during translation, which results in longer inference latency. Additionally, due to computational resource constraints, DoCIA currently only considers context from the text modality and does not include audio modality information. In the future, we plan to incorporate context from the audio modality to further enhance ST performance.
\bibliography{custom}

\appendix

\section{Prompt Templates in DoCIA}
\label{sec:prompts-shows}
\begin{figure*}[!ht]
    \centering
    \includegraphics[width=1.0\linewidth,trim=47 8 20 4,clip]{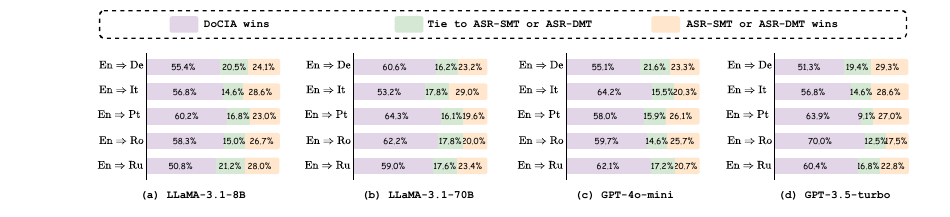}
    \caption{LLM-based evaluation results for four different base models across five ST directions. For each translation, the LLM assigns a score ranging from 1 to 10 based on the provided document-level context. \texttt{Tie to ASR-SMT} and \texttt{Tie to ASR-DMT} indicate cases where the score for DoCIA’s translation is equal to or lower than the highest score achieved by ASR-SMT and ASR-DMT, respectively.}
    \label{fig:llm-based-eval}
\end{figure*}

This section presents the prompt templates used in each stage of DoCIA. The prompt templates for ASR Refinement, machine translation and translation refinement are shown in Figure \ref{fig:asr-refine-templ}, \ref{fig:trans-templ} and \ref{fig:trans-refine-templ}, respectively. To ensure accuracy and proper formatting, we instruct the LLM to generate outputs in JSON format. 

\begin{figure*}[!t]
\begin{tcolorbox}[title=ASR Refinement Prompt Template, colback=gray!5, colframe=black, fonttitle=\bfseries]
You are an expert in automatic speech recognition refinement. Given an automatic speech recognition sentence in \texttt{<SRC-LANG>}, please check it based on its preceding automatic speech recognition sentences. Correct the capitalization, add punctuation, and eliminate incoherences such as fillers, false starts, repetitions, corrections, hesitations, and interjections. Maintain the original meaning and structure of the sentence and make it more coherent with the preceding ASR sentence. Provide your output in the following JSON format:

\{'Output': \texttt{<Refined ASR sentence>}\}
\\

\textbf{\#\#\# Preceding ASR sentences:}

\texttt{<Preceding ASR sentence>}
\\

\textbf{\#\#\# Draft current ASR sentence:}

\texttt{<Draft current ASR sentence>}
\\

\textbf{\#\#\# Your output:}
\end{tcolorbox}
\caption{Prompt template for ASR Refinement in DoCIA.}
\label{fig:asr-refine-templ}
\end{figure*}

\begin{figure*}[!t]
\begin{tcolorbox}[title=Context-aware Translation Prompt Template, colback=gray!5, colframe=black, fonttitle=\bfseries]
You are a professional translator from \texttt{<SRC-LANG>} to \texttt{<TGT-LANG>}. Given a current source sentence, please translate it to \texttt{<TGT-LANG>} based on its preceding source sentence and translation history. The translation of the current sentence should be more coherent with its preceding translations and have better lexical cohesion. Provide your translation in the following JSON format:"

\{'Output': \texttt{<Translation>}\}
\\

\textbf{\#\#\# Preceding source sentences:}

\texttt{<Preceding source sentences>}
\\

\textbf{\#\#\# Preceding translation history:}

\texttt{<Preceding translation history>}
\\

\textbf{\#\#\# Current source sentence:}

\texttt{<Current source sentence>}
\\

\textbf{\#\#\# Your output:}
\end{tcolorbox}
\caption{Prompt template for context-aware translation in DoCIA.}
\label{fig:trans-templ}
\end{figure*}

\begin{figure*}[!t]
\begin{tcolorbox}[title=Translation Refinement Prompt Template, colback=gray!5, colframe=black, fonttitle=\bfseries]
You are a professional \texttt{<SRC-LANG} to \texttt{<TGT-LANG>} translation post-editor. Given a current source sentence and its draft translation, please refine the draft translation based on its preceding source sentence and translation history. The refined translation should have the same semantics as the current source sentence be more coherent and have better lexical cohesion with its preceding translation history. Provide the refined translation in the following JSON format:
        
\{'Output': \texttt{<Refined Translation>}\}
\\

\textbf{\#\#\# Preceding source sentences:}

\texttt{<Preceding source sentences>}
\\

\textbf{\#\#\# Preceding translation history:}

\texttt{<Preceding translation history>}
\\

\textbf{\#\#\# Current source sentence:}

\texttt{<Current source sentence>}
\\

\textbf{\#\#\# Draft translation:}

\texttt{<Draft translation>}
\\

\textbf{\#\#\# Your output:}
\end{tcolorbox}
\caption{Prompt template for context-aware translation refinement in DoCIA.}
\label{fig:trans-refine-templ}
\end{figure*}

\section{GPT-4o-based Evaluation}
\label{sec:llm-base-eval-all}
Given the document-level context, source sentence, and system outputs, we ask \texttt{GPT-4o} to assign a score to each translation. The overall win rate, defined as the percentage of cases where a translation achieves the highest score, serves as a quantitative measure of document-level translation quality. As shown in Figure \ref{fig:llm-based-eval}, DoCIA achieves a notable advantage, with a win rate exceeding 57\%, outperforming both ASR-SMT and ASR-DMT. This highlights the effectiveness of DoCIA and emphasizes the importance of incorporating document-level context in speech translation tasks.

\section{Performance of ASR Refinement}
\label{sec:asr-refine-res}
\begin{figure}[!h]
\centering
\includegraphics[width=1.0\linewidth,trim=20 14 20 12,clip]{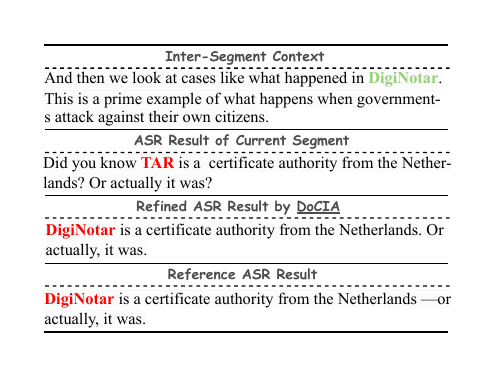}
\caption{A case study for context-aware ASR refinement. ASR result is from \texttt{Whisper-Medium}.}
\label{fig:asr-example}
\end{figure}

\begin{table}[!ht]
    \centering
    \resizebox{\linewidth}{!}{
    \begin{NiceTabular}{l|crc}
      \toprule
      {\bf Refining Model}   &  {\bf WER}$\downarrow$ &{\bf ERA} &{\bf BERTScore} 
      \\
      \midrule
      \texttt{LLaMA-3.1-8B}  & 19.16 & 55.81 & 88.75 \\
      \texttt{LLaMA-3.1-70B} & 18.66 & 56.45  &88.97\\
      \texttt{GPT-4o-mini}   & 16.01 &\underline{\bf 57.12}  &\underline{\bf 89.21}\\
     \texttt{GPT-3.5-turbo}  & 18.96 & 56.87  &89.05\\
     \midrule
     Draft ASR      &\underline{\bf 14.71} & 55.78& 88.50\\
     \bottomrule
    \end{NiceTabular}}
    \caption{Performance comparison of ASR refinement when using various LLMs.}
    \label{tab:asr-result}
\end{table}
\noindent In this section, we evaluate the performance of ASR refinement in addition to the main translation performance. Apart from Word Error Rate (WER),\footnote{In this paper, we retain the punctuation from the ASR results and report the case-sensitive WER.} we compare BERTScore \cite{Zhang2019BERTScoreET} and  Entity Recognition Accuracy (ERA) to assess how well the models utilize context to improve semantic accuracy and correct entity recognition errors. ERA is evaluated using \texttt{GPT-4o}. Specifically, we first use \texttt{GPT-4o} to extract entities from both the refined and non-refined ASR outputs (draft ASR), as well as from the reference ASR. ERA is calculated by comparing the extracted entities to the reference.

As shown in Table~\ref{tab:asr-result}, although WER increases after refinement, both ERA and BERTScore show improvements. This indicates that leveraging document-level context significantly enhances entity recognition and semantic accuracy.

Additionally, we present a case study in Figure \ref{fig:asr-example}, where DoCIA corrects an ASR error. In this case, "\textit {DigiNotar}" is misrecognized as "\textit{TAR}" in draft ASR, but DoCIA successfully corrects the error by considering the inter-segment context, which include the proper entity "\textit{DigiNotar}".

\section{Accuracy of Pronoun Translation}
\label{sec:apt-eval}
\begin{table}[!ht]
    \centering
    \resizebox{0.9\linewidth}{!}{
    \begin{NiceTabular}{l|cccc}
    \toprule
    \Block[l]{2-1}{\textbf{System}} & \Block[c]{1-2}{\textbf{En $\Rightarrow$ De}} &  & \Block[c]{1-2}{\textbf{En $\Rightarrow$ Ru}}\\
    
    \cmidrule(lr){2-3}\cmidrule(lr){4-5}
    & APT & LTCR & APT & LTCR \\
    
    \rowcolor[gray]{0.8}
    \Block[c]{1-5}{\texttt{LLaMA-3.1-8B}}\\
    ASR-SMT                    & 55.34 & 53.20 & 54.88 & 52.88\\
    ASR-DMT                    & 56.15 & 53.80 & 55.51 & 53.69\\
    \hdashedline
    DoCIA                      &\underline{\bf 58.33} &\underline{\bf 55.53} &\underline{\bf 56.99} &\underline{\bf 54.77}\\
    
    \rowcolor[gray]{0.8}
    \Block[c]{1-5}{\texttt{GPT-4o-mini}}\\
    ASR-SMT                    & 59.15 & 56.33 & 58.97 & 55.39\\
    ASR-DMT                    & 59.89 & 57.01 & 59.56 & 56.11\\
    \hdashedline
    DoCIA                      &\underline{\bf 61.92} &\underline{\bf 58.21} &\underline{\bf 62.11} &\underline{\bf 57.01} \\
    \bottomrule
    \end{NiceTabular}}
   
    \caption{Performance comparison for Accuracy of Pronoun Translation (APT) and Lexical Translation Consistency Rate (LTCR).}
\label{tab:apt-ltcr}
\end{table}
\noindent We evaluate the accuracy of pronoun translation (APT), a reference-based metric introduced by~\citet{miculicich-werlen-popescu-belis-2017-validation}. As shown in Table \ref{tab:apt-ltcr}, DoCIA demonstrates an improvement in pronoun translation performance over both ASR-SMT and ASR-DMT, suggesting DoCIA is effective in addressing coreference and anaphora in ST task.

\section{Lexical Translation Consistency}
\label{sec:ltcr-eval}
In addition to evaluating pronoun translation accuracy, we also assess lexical translation consistency rate (LTCR), a reference-free metric introduced by \citet{lyu_etal_2021_emnlp_ltcr}, to measure whether the same word is consistently translated into the target language. As shown in Table \ref{tab:apt-ltcr}, DoCIA also significantly outperforms both ASR-SMT and ASR-DMT.

\section{Effect of ASR Model}
\label{sec:asr-effect}
\begin{table}[!ht]
    \centering
    \resizebox{\linewidth}{!}{
    \begin{NiceTabular}{l|cccc}
    \toprule
    \Block[l]{2-1}{\textbf{System}} & \Block[c]{1-2}{\textbf{En $\Rightarrow$ De}} &  & \Block[c]{1-2}{\textbf{En $\Rightarrow$ Ru}}\\
    
    \cmidrule(lr){2-3}\cmidrule(lr){4-5}
    & $s$-Comet & $d$-Comet & $s$-Comet & $d$-Comet \\
    
    \rowcolor[gray]{0.8}
    \Block[c]{1-5}{\texttt{LLaMA-3.1-8B}}\\
    ASR-SMT                    & 78.01 & 5.680 & 76.36 & 5.168\\
    ASR-DMT                    & 77.88 & 5.712 & 76.99 & 5.211\\
    \hdashedline
    DoCIA (\textit{w/ WM})                   & \underline{\bf 79.15} & 5.912 & 78.39 & 5.556\\
    ~~\textit{w/ WS, WER=14.89} & 78.99 & 5.901 & 78.45 & \underline{\bf 5.562}\\
    ~~\textit{w/ WL, WER=14.41} & 79.11 & \underline{\bf 5.935} & \underline{\bf 78.61} & 5.551 \\
    
    \rowcolor[gray]{0.8}
    \Block[c]{1-5}{\texttt{GPT-4o-mini}}\\
    ASR-SMT                    & 82.01 & 6.001 & 82.21 & 5.827\\
    ASR-DMT                    & 82.42 & 6.108 & 82.80 & 5.948\\
    \hdashedline
    DoCIA (\textit{w/ WM})             & 83.64 & 6.444 & 84.32 & 6.286\\
    ~~\textit{w/ WS, WER=14.89}  & 83.43 & 6.401 & 84.34 & 6.275\\
    ~~\textit{w/ WL, WER=14.41} & \underline{\bf 83.82} & \underline{\bf 6.478} & \underline{\bf 84.71} & \underline{\bf 6.299} \\
    \bottomrule
    \end{NiceTabular}}
   
    \caption{Performance comparison when using various ASR models. \textbf{WS}, \textbf{WM} and \textbf{WL} denote the \texttt{Whisper-Small}, \texttt{Whisper-Medium} and \texttt{Whisper-Large} models, respectively.}
\label{tab:asr-model-effect}
\end{table}

In our experiments, DoCIA uses the \texttt{Whisper-Medium} ASR model to generate segment-level transcriptions. We now investigate the effect of using ASR models of different sizes on the final translation performance. Table \ref{tab:asr-model-effect} presents a comparison of translation results across different ASR models. It shows that larger ASR models tend to achieve better ASR performance (i.e., lower WER), leading to modest improvements in translation quality. For instance, using the \texttt{Whisper-Large} yields a +0.39 improvement in the $s$-COMET score for the En$\Rightarrow$Ru task compared to the \texttt{Whisper-Medium}, when DoCIA uses the \texttt{GPT-4o-mini} translation model.

\section{Effect of Hyperparameter $\lambda$ in Refinement Determination}
\label{sec:lamda-func}
\begin{table}[!t]
    \centering
    \resizebox{\linewidth}{!}{
    \begin{NiceTabular}{c|cccc}
    \toprule
    \Block[l]{2-1}{\textbf{System}} & \Block[c]{1-2}{\textbf{En $\Rightarrow$ De}} &  & \Block[c]{1-2}{\textbf{En $\Rightarrow$ Ru}}\\
    
    \cmidrule(lr){2-3}\cmidrule(lr){4-5}
    & $s$-Comet & $d$-Comet & $s$-Comet & $d$-Comet \\
    
    \rowcolor[gray]{0.8}
    \Block[c]{1-5}{\texttt{LLaMA-3.1-8B}}\\
    
    DoCIA ($\lambda=0.7$) & \underline{\bf 79.15} & \underline{\bf 5.912} & 78.39 & \underline{\bf 5.556}\\
    \hdashedline
    ~~~~$\lambda=0.0$ & 78.24 & 5.735 & 77.56 & 5.441 \\
    ~~~~$\lambda=0.5$   & 78.54 & 5.733 & 77.81 & 5.533\\
    ~~~~$\lambda=0.9$ & 78.21 & 5.712 & 77.31 & 5.432 \\
    ~~~~$\lambda=1.0$ & 78.33 & 5.812 & 77.50 & 5.431 \\
    \hdashedline
    
    ~~~~\textit{w/} BS ($\lambda=0.7$) & 78.81 & 5.865 & \underline{\bf 78.45} & 5.511\\
    
    \rowcolor[gray]{0.8}
    \Block[c]{1-5}{\texttt{GPT-4o-mini}}\\
     DoCIA ($\lambda=0.7$)              & 83.64 & \underline{\bf 6.444} & \underline{\bf 84.32} & \underline{\bf 6.286}\\
    \hdashedline
    ~~~~$\lambda=0.0$   & 82.79 & 6.259 & 83.33 & 6.199\\
    ~~~~$\lambda=0.5$   & 83.31 & 6.387 & 83.83 & 6.218\\
    ~~~~$\lambda=0.9$   & 82.98 & 6.253 & 83.45 & 6.166\\
    ~~~~$\lambda=1.0$  & 82.66 & 6.299 & 83.11 & 6.116\\
    
    \hdashedline
    ~~~~\textit{w/} BS ($\lambda=0.7$) & \underline{\bf 83.75} & 6.393 & 84.12 & 6.201\\
    \bottomrule
    \end{NiceTabular}}
   
    \caption{Performance comparison when setting different $\lambda$. When setting $\lambda=1.0$ (or $\lambda=0.0$), we always take the original (or refined) text as the final output.}
\label{tab:llama-func-effect}
\end{table}

To prevent hallucinations in both the transcription and translation refinement processes, we introduce a refinement determination mechanism. In this section, we investigate the impact of the determination threshold, $\lambda$, and explore the effect of using BERTScore to compute the similarity between $I$ (input) and $O$ (output) by replacing Eq. \ref{equ:idt-f} with BERTScore. The results, presented in Table \ref{tab:llama-func-effect}, show that both excessively high and low threshold values negatively affect performance. Additionally, using BERTScore in the refinement determination process leads to significant performance improvements. This suggests that the determination mechanism is not highly sensitive to the choice of similarity function.

\section{Details of Human Evaluation}
\label{sec:details-da}
\paragraph{Recruitment and Criterion.} We recruit evaluators who are professional translators with a minimum of five years of experience. Given a reference ASR output, its translations from various systems, and the human-produced reference translation, evaluators are tasked with assigning a score on a scale from 0 to 100. The detailed scoring criterion as follows:
\begin{itemize}
    \item \texttt{0-20}: The translation is completely incorrect and unclear, with only a few words or phrases being correct. It is totally unreadable and difficult to understand.
    \item \texttt{21-40}: The translation has very little semantic similarity to the source sentence, with key information missing or incorrect. It has numerous unnatural and unfluent expressions and grammatical errors.
    \item \texttt{41-60}: The translation can express part of the key semantics but has many non-key semantic errors. It lacks fluency and idiomaticity.
    \item \texttt{61-80}: The translation can express the key semantics but has some non-key information errors and significant grammatical errors. It lacks idiomaticity.
    \item \texttt{81-100}: The translation can express the semantics of the source sentence with only a few non-key information errors and minor grammatical errors. It is fluent and idiomatic.
\end{itemize}

\paragraph{Coherence Error and Content Error.} We manually count the average number of coherence errors (CE) and content translation errors (CTE) for evaluation terms. Specifically, CE involves two types of errors, including inter-sentential consistency errors, such as inconsistent translations of the same entity across sentences, and inter-sentential logical errors, such as improper translation or usage of transition words and conjunctions. CTE includes three error types: mistranslation, under- and over-translation.

\end{document}